\documentclass[conference]{IEEEtran}
\IEEEoverridecommandlockouts
\usepackage{cite}
\usepackage{amsmath,amssymb,amsfonts}
\usepackage{algorithmic}
\usepackage{graphicx}
\usepackage{textcomp}
\usepackage{xcolor}
\usepackage{etoolbox}
\def\BibTeX{{\rm B\kern-.05em{\sc i\kern-.025em b}\kern-.08em
    T\kern-.1667em\lower.7ex\hbox{E}\kern-.125emX}}
\begin{document}

\title{Improving Clinical Efficiency and Reducing Medical Errors through NLP-enabled diagnosis of Health Conditions from Transcription Reports\\}

\author{
  Krish Maniar\textsuperscript{*}\thanks{* Equal Contribution}\\
  \texttt{krishmaniar4@gmail.com}
  \and
  Shafin Haque\textsuperscript{*}\\
  \texttt{shafin1025@gmail.com}
  \and
  Kabir Ramzan\textsuperscript{*}\\
  \texttt{kabir@ramzan.me}
}

\maketitle

\begin{abstract} 
Misdiagnosis rates are one of the leading causes of medical errors in hospitals, affecting over 12 million adults across the US. To address the high rate of misdiagnosis, this study utilizes 4 NLP-based algorithms to determine the appropriate health condition based on an unstructured transcription report. From the Logistic Regression, Random Forest, LSTM, and CNN-LSTM models, the CNN-LSTM model performed the best with an accuracy of 97.89\%. We packaged this model into a authenticated web platform for accessible assistance to clinicians. Overall, by standardizing health care diagnosis and structuring transcription reports, our NLP platform drastically improves the clinical efficiency and accuracy of hospitals worldwide. \footnote{Code available at \texttt{https://github.com/CMEONE/MedicAI}}
\end{abstract}

\begin{IEEEkeywords}
Natural Language Processing, Medical Transcription Notes(s), Diagnostic Systems, Recurrent Neural Network, Long Short-Term Memory Network
\end{IEEEkeywords}


\section{Introduction}

\subsection{\textbf{Physician Ineffectiveness in Hospitals}}

In the United States alone, nearly 400,000 people are killed annually due to preventable medical errors \cite{b1}. With over 12 million adults in the U.S. receiving misdiagnoses yearly, inaccurate diagnoses are the leading cause of medical errors in hospitals \cite{b2}. Misdiagnoses refer to inaccurate assessments by healthcare providers of a patient's condition, often leading to either inappropriate or excessive treatment and sometimes no treatment at all. In a case study conducted at the University of Maryland Medical Center, it was found that in a group of 177 patients, over 90\% of patients received at least one unnecessary treatment, highlighting the unreliable nature of physicians in hospital settings \cite{b3}. Not only can misdiagnoses undermine the effectiveness of clinicians, but over one-third of such cases with inaccurate treatment result in death, injury, or permanent disabilities \cite{b4}.

Beyond high misdiagnosis rates, the lack of communication between patients and doctors contributes to diminished patient outcomes and care. In 2010, a study conducted in JAMA Internal Medicine discovered the striking discrepancies in communication between patients and doctors, especially in relation to diagnoses. The authors determined that while 77\% of physicians believed patients knew their diagnosis, only 57\% of patients actually did \cite{b5}. This alarming inconsistency reveals broader issues regarding the subjectivity, unreliability, and variability between clinicians, demanding the necessity for more objective and accurate diagnostic approaches.

\subsection{\textbf{Electronic Health Records and Unstructured Data}}

The recent rise in electronic health records (EHRs) only increases this variability and further warrants the need for automated methods to standardize diagnosis procedures in medical settings \cite{b6}. EHRs primarily consist of unstructured data, which lacks standardized specifications and is written in free-text without, clear separations. Unstructured data is often found in clinical report notes, discharge summaries, and chart reviews; however, interpreting such data requires costly time and presents a considerable challenge to medical practitioners due to its lack of baselines and standardized components. In fact, the The Agency for Healthcare Research and Quality acknowledges unstructured data in EHRs as a primary barrier in understanding data quality and interpreting key diagnostic information \cite{b7}. Currently, most teams and researchers that focus on medical record abstraction target structured data, ignoring the widespread amount of unstructured data. 

Because of the sheer amount of unstructured data available, the use of Machine Learning and Deep Learning methods are crucial for eliminating clinician subjectivity across hospitals around the world and for utilizing the vast amount of data available to improve diagnosis. \\

\section{Literature Review}

\subsection{\textbf{Emergence of Machine Learning and NLP}}

Machine Learning, a subset of Artificial Intelligence, refers to algorithms that can efficiently learn, generalize, and thus develop a complex, mathematical understanding of pre-existing training data to give accurate predictions on unseen test data \cite{b8}. As a result of the increase in computational power and the emergence of the “big data” era, the application of ML in medical imaging and health care systems has increased and can strongly facilitate healthcare operations \cite{b9}. More specifically, Natural Language Processing (NLP), a form of Deep Learning, has emerged as a promising discipline to analyze unstructured data, a task that is time-consuming and extremely challenging for doctors and physicians, especially in low-income countries. 

Over 80\% of all healthcare data is unstructured \cite{b10}, so the ability of NLP models to decode unstructured clinical reports and provide insightful diagnostic advice presents an enormous opportunity. However, NLP is most often associated with speech analytics, sentiment analysis, and chat generation, with relatively little attention for misdiagnosis rates in medical settings. As the healthcare community looks to incorporate exponentially more automated technologies into their infrastructure, it is vital to approach specific problems in the medical community with state-of-the-art NLP models. 
 
\subsection{\textbf{Objective}}

While recent studies have used models such as RNNs \cite{b11} to predict surgery complications and extract structured data from clinical reports, very few institutions have focused their research on correcting misdiagnosis rates. Moreover, there is a lack of an easily accessible and accurate early-diagnostics system for classifying medical specialties from unstructured medical transcription notes.

Thus, in an effort to streamline the process for medical practitioners and improve clinical care for patients within the hospital, we developed various machine learning and deep learning-based architectures to determine the specific medical specialty that a clinical transcription note describes, ultimately reducing misdiagnosis rates and alleviating the burden for overloaded hospitals. Furthermore, to increase accessibility for hospitals worldwide, our models are packaged into an efficient and accurate AI-enabled medical app.  \\

\section{Methods}

To categorize the medical transcription reports into various specialties and types, our machine learning models require large amounts of categorical data to operate smoothly. In addition to scraping data from an online website to curate our dataset, we also implemented various types of data preprocessing to maximize the performance of all models used.

\subsection{\textbf{Data Collection and Transformation}}

We retrieved and scraped our data from MTSamples, a public corpus of data that contains transcribed medical reports. 5003 sample reports were extracted from the MTSamples collection of data. It contains 40 different medical specialties, such as 'Urology' and 'Nephrology,' 28581 unique words, and 487.5 average characters per sentence.

\begin{figure}[h!]
  \centering
  \includegraphics[width=0.5\textwidth]{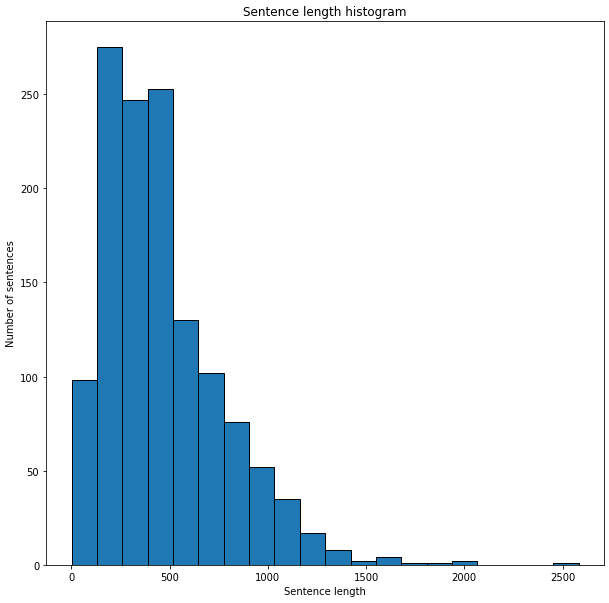}
  \caption{Transcription Report Sentence Length Histogram}
  \label{fig:sent}
\end{figure}

To reduce heterogeneity between values in the dataset, we funneled the data down to four main human body systems: Heart, Brain, Reproductive, and Digestive. To maintain a fair balance of nearly 300 reports per category, we accordingly combined certain medical specialties. For example, the Brain label contained both "Neurology" and "Neurosurgery" reports, and the Digestive label contained both "Gastroenterology" and "Nephrology." In total, our modified dataset had 1304 reports (371 heart, 317 brain, 311 reproductive, and 305 digestive), a unique vocabulary size of 20127, and a mean sentence length of 464.9 characters. A sentence length histogram is displayed in Figure \ref{fig:sent} to illustrate the distribution of sentence length for the medical transcription reports included in our modified dataset.

Within our dataset, each report contains a variety of non-standardized labels with sections differing even within medical specialties. Our scraper attempted to extract all relevant information regarding the medical specialty and transcription report details. To further resolve any discrepancies and correct errors from our scraping algorithm, we also screened the full modified dataset.

\subsection{\textbf{Data Preprocessing}}

After transforming the dataset and parsing the specified categories, we applied four preprocessing steps on cleaning the data. First, we removed any reports with missing transcription details or medical specialties. Second, we removed numbers, symbols, punctuation, and special characters such as brackets and slashes to standardize the text and delete unnecessary elements, allowing for improved analysis of the transcription input. Third, we tokenized our transcription sentences into smaller tokens, all in lower-case, while also removing stopwords that lack contextual significance. Fourth, we lemmatized our data by converting the tokenized elements into shorter but more useful forms and linking words to its root through a dictionary-based implementation. These four steps work together to convert the initially unstructured transcription report to a more standardized and structured report. 

Once our data was cleaned, we applied a recently developed vectorization preprocessing algorithm called Term Frequency Inverse Document Frequency (TFIDF) to translate the string tokens to numerical vectors. Most natural language processing models use the Bag Of Words (BOW) algorithm, where the location of words directly influences the text vectorization. However, this approach fails to utilize semantics to guide the vectorization of an embedding space. The TFIDF algorithm builds upon this traditional BOW approach and consists of two primary steps: Term Frequency and Inverse Data Frequence. The first step divides the frequency of each word by the total number of words in the text. The second step weights each word based on its semantic relevance to the medical specialty. For example, words like "the" and "of" may be the most frequent tokens but lack semantic value to the classification task, so they will be assigned a low weight. Both steps work together to produce a vector and represent the reports in an embedding space. Figure \ref{fig:sent} displays the embedding space for all four categories in the dataset (Brain, Heart, Digestive, and Reproductive). 

\begin{figure}[h!]
    \centering
  \includegraphics[width=0.4\textwidth]{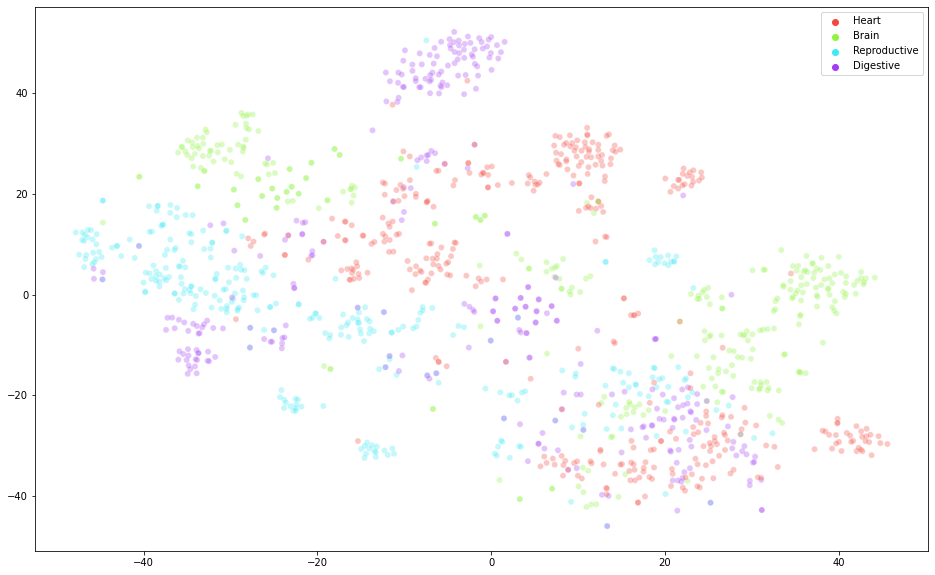}
  \caption{TFIDF Vector Embeddings for Modified Dataset}
  \label{fig:tfidf}
\end{figure}

While the embeddings are segregated in certain portions of the figure, they overlap in numerous areas, which both explains the discrepancies that doctors have in classifying such reports and adds to the complexity of our data and models.

\subsection{\textbf{Machine Learning Models}}

Once we processed our data into an appropriate format, we developed 2 ML and 2 DL models. All models were trained and tested with a dataset consisting of an 80\% train and a 20\% test split.

\begin{figure}[h!]
    \centering
  \includegraphics[width=0.45\textwidth]{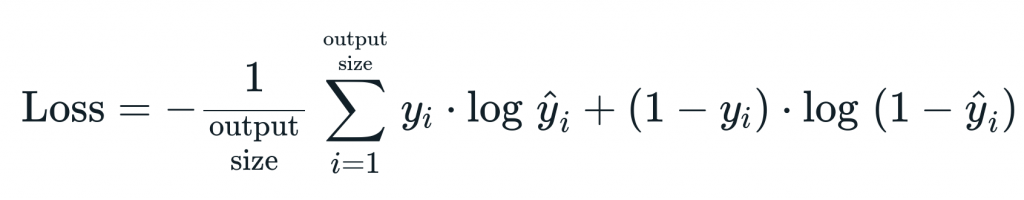}
  \caption{Category Cross Entropy Loss Function}
  \label{fig:catentr}
\end{figure}

\textbf{Logistic Regression}. Our first ML algorithm (Logistic Regression) generally uses a logistic function to separate two classes from each other through using a binary entropy function. However, our dataset contains multiple classes, so we modified our loss function from binary to categorical class entropy, a multinomial probability distribution function that computes the loss over all four classes. The loss function formula is displayed in Figure \ref{fig:catentr}. 

We also applied Principal Component Analysis with 95\% variance to reduce the dimensionality of the features being inputted into the model. This helped expedite the training process and improve the interpretability of our Logistic Regression model.

\begin{figure}[h!]
    \centering
  \includegraphics[width=0.4\textwidth]{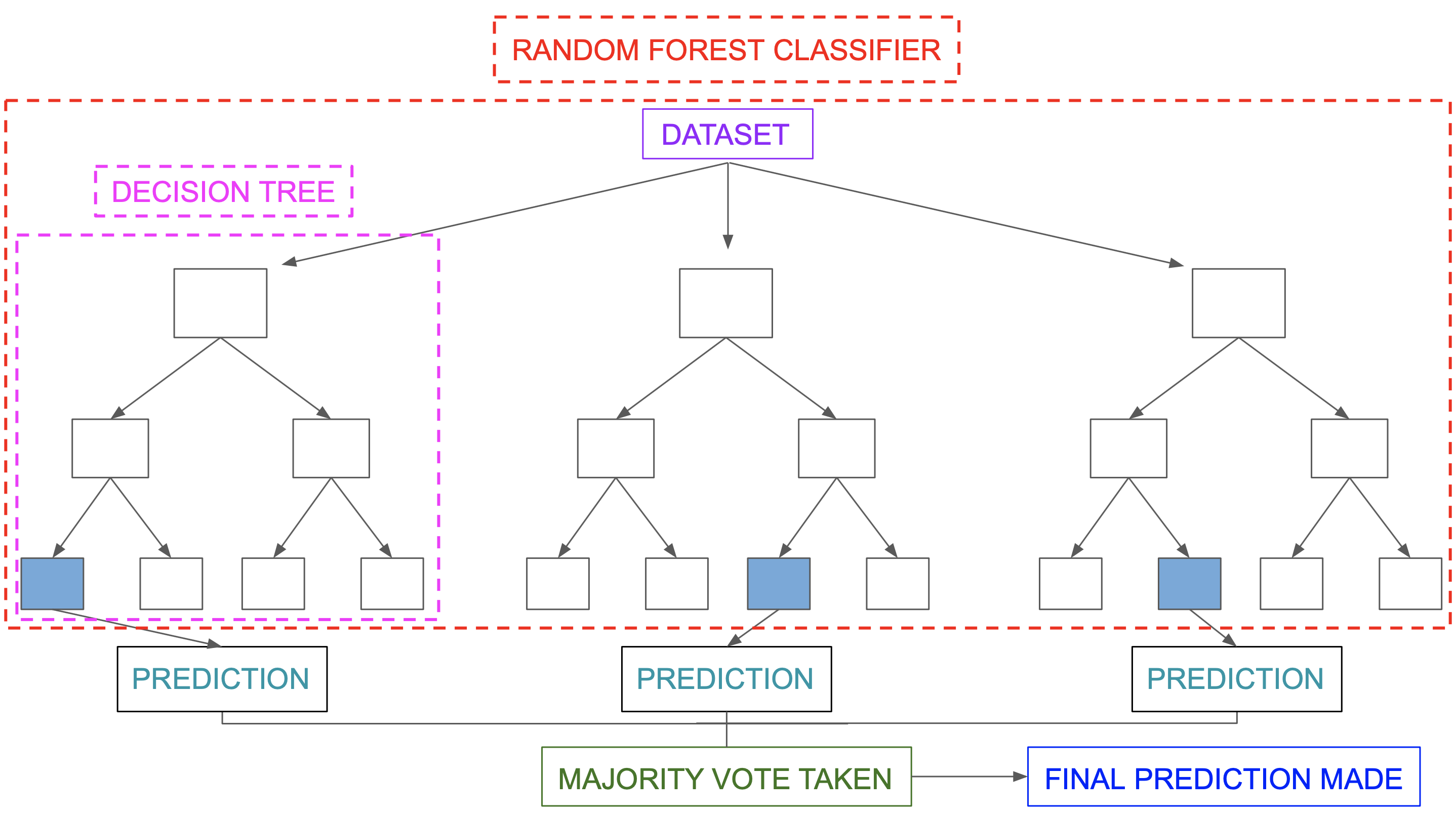}
  \caption{Random Forest Architecture \cite{b12}}
  \label{fig:rf}
\end{figure}

\textbf{Random Forest}. In addition to Logistic Regression, we used Random Forest, another common Machine Learning architecture. As shown in Figure \ref{fig:rf}, this model consists of an ensemble of decision trees and uses numerous yes/no branches to deliver a result for classification problems. After fine-tuning several hyper parameters, namely the number of decision trees and respective branch depth, we determined the optimal amount of estimators was 150 trees and ideal maximum amount of depth was 4 branches.

\subsection{\textbf{Deep Learning Models}}

ML models such as Logistic Regression lack the ability to interpret semantics and maintain long-term connections in text. Thus, we also developed several Deep Learning methods: an LSTM and a CNN-LSTM, which combine an LSTM's semantic elements and a CNN's spatial dependencies.

\begin{figure}[h!]
  \centering
  \includegraphics[width=0.4\textwidth]{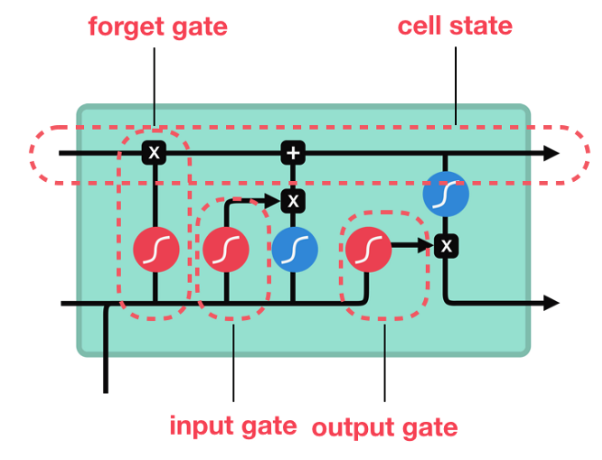}
  \caption{LSTM memory cell and corresponding gates \cite{b13}}
  \label{fig:lstm}
\end{figure}

\textbf{Long Short-Term Memory Network}. By constructing long-term connections in the transcription reports, the LSTM model is an improvement over Recurrent Neural Networks. A key difference between LSTM and RNN architectures is the presence of a memory cell that maintains a hidden state as it progresses through the text. Figure \ref{fig:lstm} illustrates the details of a memory cell. The states are regulated by the forget, input gate, and output gate, which alter memory connections between segments of the data.

\begin{figure}[h!]
  \centering
  \includegraphics[width=0.45\textwidth]{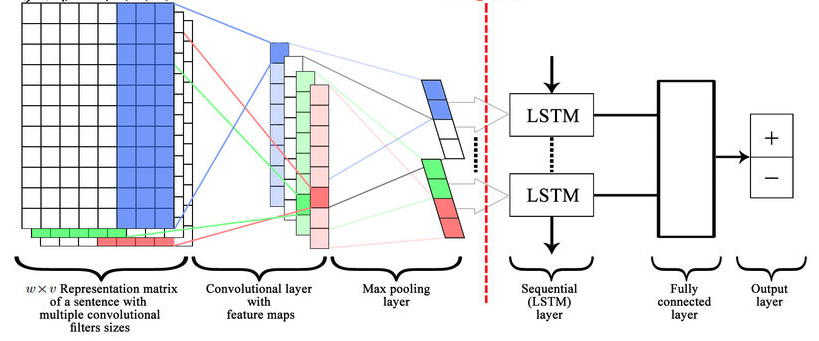}
  \caption{CNN-LSTM architecture\cite{b14}}
  \label{fig:cnnlstm}
\end{figure}

\textbf{CNN-LSTM}. While CNN models are generally used for computer vision tasks, they can provide useful information from text if used in conjunction with an LSTM model. Because CNNs use convolutional layers and padding, they can extract higher-level features than LSTMs. However, they are unable to track long-term dependencies in transcription reports. Thus, a combination of CNN and LSTM models would be expected to outperform a core RNN or LSTM model. Similar training parameters were utilized as the LSTM network, and an overview of the model architecture is visible in Figure \ref{fig:cnnlstm}.

\subsection{\textbf{MedicAI Web Application}}

\begin{figure}[h!]
  \centering
  \includegraphics[width=0.45\textwidth]{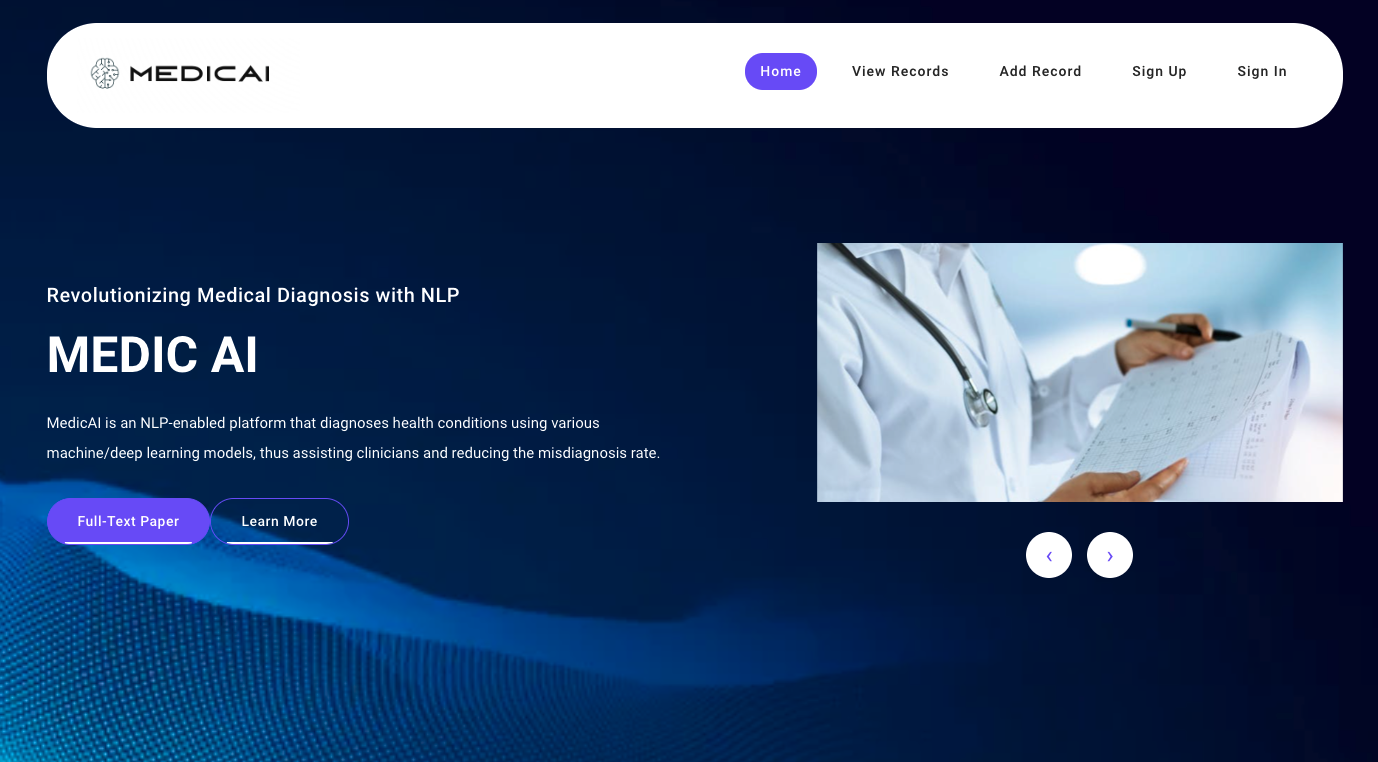}
  \caption{Dashboard for MedicAI Web Application}
  \label{fig:dash}
\end{figure}

To package our Machine and Deep Learning models into an accessible platform, we created a Flask web application called MedicAI that serves as a dashboard for clinicians. As shown in the dashboard in Figure \ref{fig:dash}, doctors are able to access medical transcript records for all their patients, and they can accordingly add and filter patient records. Once a record is uploaded as an image, our Optical Character Recognition (OCR) algorithm will tokenize the text from the image and parse the corresponding "Findings" section of the report. Then, as described in the 'Data Preprocessing' section, this parsed text will be preprocessed using TFIDF and subsequently inputted to the model for diagnosis classification. 

In terms of the web technologies we used, our front-end was made with HTML, CSS, and JavaScript, and the back-end infrastructure was developed with Flask and Firebase. We implemented authentication procedures with secure login/logout functionality using Firebase and developed the database using Firebase Cloud Firestore. We believe our application is crucial for assisting clinicians in health condition diagnoses, and it has the potential of leveling the playing field between hospitals of different areas.

\section{Results}

\begin{figure}[h!]
  \centering
  \includegraphics[width=0.35\textwidth]{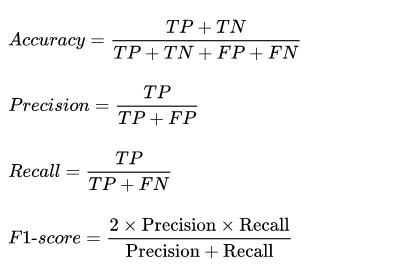}
  \caption{Formulas for Performance Metrics}
  \label{fig:metrics}
\end{figure}

Once we developed Logistic Regression, Random Forest, LSTM, and CNN-LSTM, we trained the models on 80\% of the dataset and tested them on the remaining 20 percent. The Machine Learning models were developed with SciKit-Learn, and the Deep Learning models were developed with TensorFlow. All models were trained with a categorical cross-entropy loss, a softmax activation function, and 50 epochs. To analyze the performance of each model, we calculated 4 commonly used performance metrics: Accuracy, Precision, Recall, and F1-score. Figure \ref{fig:metrics} displays a detailed breakdown of the formulas for each performance metric based on True Positive (TP), True Negative (TN), False Positive (FP), and False Negative (FN) rates.

After tuning the Logistic Regression model's hyperparameters with a gridsearch function, the model yielded an accuracy of 91.19\% accuracy. Table I describes the results for the other performance metrics.

\begin{table}[htbp]
\caption{Logistic Regression performance on the Test Dataset}
  \begin{center}
    
    \label{tab:table1}
    \begin{tabular}{c|c|c|c}
      & \textbf{Precision} & \textbf{Recall} & \textbf{F1-score} \\
      \hline
      Heart & 0.92 & 0.95 & 0.93\\
      Brain & 0.94 & 0.94 & 0.94\\
      Reproductive & 0.90 & 0.89 & 0.89\\
      Digestive & 0.88 & 0.87 & 0.88\\ 
      \end{tabular}
    
  \end{center}
\end{table}

As for the Random Forest model, the resulting accuracy on the test set was 84.29\% accuracy. Other performance metrics are displayed in Table II. The optimal hyperparameters were 150 decision trees, each with a depth of 4 branches. The other performance metrics were also computed in Table II.

\begin{table}[htbp]
\caption{Random Forest performance on the Test Dataset}
  \begin{center}
    
    \label{tab:table2}
    \begin{tabular}{c|c|c|c}
      & \textbf{Precision} & \textbf{Recall} & \textbf{F1-score} \\
      \hline
      Heart & 0.81 & 0.91 & 0.85\\
      Brain & 0.89 & 0.92 & 0.91\\
      Reproductive & 0.86 & 0.77 & 0.79\\
      Digestive & 0.82 & 0.75 & 0.79\\ 
      \end{tabular}
    
  \end{center}
\end{table}

For most of the metrics, the Logistic Regression and Random Forest models performed the best for the Brain class, with lower performance on the Reproductive and Digestive classes. Additionally, the more traditional Logistic Regression performed with an accuracy of nearly 7\% higher, suggesting its simplicity outperforms the complex ensembled decision tree estimators.

While the two Machine Learning models performed at an acceptable accuracy, the Deep Learning methods——LSTM and CNN-LSTM——produced much higher performance rates. The Deep Learning models were both trained for 50 epochs on a 20\% train set with a batch size of 100. The LSTM model performed at an accuracy of 97.60\%, and the CNN-LSTM performed slightly better with an accuracy of 97.89\%. More detailed performance metrics are illustrated in Table III and Table IV below for the LSTM and CNN-LSTM model, respectively.

\begin{table}[htbp]
\caption{LSTM performance on the Test Dataset}
  \begin{center}
    \label{tab:table3}
    \begin{tabular}{c|c|c|c}
      & \textbf{Precision} & \textbf{Recall} & \textbf{F1-score} \\
      \hline
      Heart & 1.00 & 0.97 & 0.98\\
      Brain & 0.98 & 1.00 & 0.99\\
      Reproductive & 0.97 & 0.98 & 0.97\\
      Digestive & 0.96 & 0.96 & 0.96\\ 
      \end{tabular}
  \end{center}
\end{table}

\begin{table}[htbp]
\caption{CNN-LSTM performance on the Test Dataset}
  \begin{center}
    \label{tab:table4}
    \begin{tabular}{c|c|c|c}
      & \textbf{Precision} & \textbf{Recall} & \textbf{F1-score} \\
      \hline
      Heart & 1.00 & 0.97 & 0.98\\
      Brain & 0.99 & 1.00 & 0.99\\
      Reproductive & 1.00 & 0.96 & 0.98\\
      Digestive & 0.93 & 0.99 & 0.96\\ 
      \end{tabular}
  \end{center}
\end{table}

Overall, the CNN-LSTM performed the best for most performance metrics, even outperforming the standard LSTM model. This is likely due to the spatial features of the CNN that facilitate the semantical and long-term memory connections computed by the LSTM model.

\section{Discussion}

In this study, we developed a web application for doctors that utilizes both traditional and modern Machine Learning methods to improve diagnosis of health conditions from medical transcription notes. Not only does this app reduce misdiagnosis rates and increase clinical efficiency in hospitals, but it also standardizes medical diagnosis across hospitals around the world with an affordable and accurate tool. 

After constructing a Logistic Regression, Random Forest, LSTM, and CNN-LSTM model, we determined that the CNN-LSTM model, which incorporates spatial features and long-term memory connections, performed the best with an accuracy of 97.89\% and an average F1-score of 0.98. Our model was able to seamlessly correctly classify Brain, Heart, Reproductive, and Digestive categories. We integrated this CNN-LSTM model into our platform called MedicAI to classify health conditions in real-time based on medical transcription notes, satisfying the goals of this project.

Further research into state-of-the-art NLP models and preprocessing methods should be explored. In the future, we should look into modern NLP models, such as BERT and GPT, which utilize transformer architectures and self-supervised learning to learn improved representations. To support the computational complexity brought upon by these models, we also hope to develop our back-end infrastructure to support larger ML models. Additionally, we plan to test varying preprocessing methods for converting tokens into vectors in an embedding space, such as Glove and Word2Vec. Lastly, we hope to expand the available data and scope of the project by incorporating institutional data from hospitals around the world, and eventually implementing this platform in global medical centers. Thus, we hope this project serves as a strong basis for future research into the use of advanced NLP methods for medical specialty classification from transcription notes.

\end{document}